\documentclass[10pt,twocolumn,letterpaper]{article}
\usepackage{iccv,times,graphicx,amsmath,amssymb,booktabs,tabulary,multirow,overpic,xcolor,algorithm,algorithmic,array,float,subfig}

\newcolumntype{I}{!{\vrule width 3pt}}
\newlength\savedwidth

\newlength\savewidth
\newcommand\shline{\noalign{\global\savewidth\arrayrulewidth
 \global\arrayrulewidth 1pt}%
 \hline
\noalign{\global\arrayrulewidth\savewidth}}

\newcolumntype{x}[1]{>{\centering\arraybackslash}p{#1pt}}

\newcommand{\tablestyle}[2]{\setlength{\tabcolsep}{#1}\renewcommand{\arraystretch}{#2}\centering\footnotesize}
\makeatletter\renewcommand\paragraph{\@startsection{paragraph}{4}{\z@}
{.3em \@plus1ex \@minus.2ex}{-.5em}{\normalfont\normalsize\bfseries}}\makeatother

\iccvfinalcopy

\usepackage[pagebackref=true,breaklinks=true,letterpaper=true,colorlinks,bookmarks=false]{hyperref}

\def\iccvPaperID{*} 

\ificcvfinal\fi

\begin{document}

\title{ISTR: End-to-End Instance Segmentation with Transformers}

\author{
Jie Hu$^1$, Liujuan Cao$^1$, Yao Lu$^1$, ShengChuan Zhang$^1$, Yan Wang$^2$, \\Ke Li$^3$, Feiyue Huang$^3$, Ling Shao$^4$, and Rongrong Ji$^{1,5}$.\\
$^1$Media Analytics and Computing Lab, Department of Artificial Intelligence,\\ School of Informatics, Xiamen University,
$^2$Pinterest,
$^3$Tencent Youtu Lab, \\
$^4$Inception Institute of Artificial Intelligence,
$^5$Institute of Artificial Intelligence, Xiamen University.
}

\maketitle
\ificcvfinal\thispagestyle{empty}\fi

\begin{abstract}
End-to-end paradigms significantly improve the accuracy of various deep-learning-based computer vision models.
To this end, tasks like object detection have been upgraded by replacing non-end-to-end components, such as removing non-maximum suppression by training with a set loss based on bipartite matching.
However, such an upgrade is not applicable to instance segmentation, due to its significantly higher output dimensions compared to object detection.
In this paper, we propose an instance segmentation Transformer, termed ISTR, which is the first end-to-end framework of its kind.
ISTR predicts low-dimensional mask embeddings, and matches them with ground truth mask embeddings for the set loss.
Besides, ISTR concurrently conducts detection and segmentation with a recurrent refinement strategy, which provides a new way to achieve instance segmentation compared to the existing top-down and bottom-up frameworks.
Benefiting from the proposed end-to-end mechanism, ISTR demonstrates state-of-the-art performance even with approximation-based suboptimal embeddings.
Specifically, ISTR obtains a 46.8/38.6 box/mask AP using ResNet50-FPN, and a 48.1/39.9 box/mask AP using ResNet101-FPN, on the MS COCO dataset.
Quantitative and qualitative results reveal the promising potential of ISTR as a solid baseline for instance-level recognition.
Code has been made available at: \url{https://github.com/hujiecpp/ISTR}.
\end{abstract}

\section{Introduction}
A growing trend in the recent development of computer vision is to remove the handcrafted components to enable end-to-end training and inference, which has demonstrated significant improvement in multiple fields.
However, this end-to-end paradigm still lacks applications for instance segmentation that aims to jointly detect and segment each object in an image.
Existing instance segmentation approaches either need a manually-designed post-processing step called non-maximum suppression (NMS) to remove duplicate predictions~\cite{he2017mask,liu2018path,huang2019mask,chen2020blendmask,lee2020centermask,wang2020solo}, or are early trials on small datasets and lack evaluation against modern baselines~\cite{ren2017end,romera2016recurrent}.
Popular approaches also rely on a top-down or bottom-up framework that decomposes instance segmentation into several dependent tasks, preventing them from being end-to-end.

Besides instance segmentation, object detection also faces similar challenges.
Recent studies enable end-to-end object detection by introducing a set prediction loss~\cite{hu2018relation,carion2020end,sun2020sparse,wang2020end,zhu2020deformable}, with optional use of Transformers~\cite{vaswani2017attention}.
The set prediction loss enforces bipartite matching between labels and predictions to penalize redundant outputs, thus avoiding NMS during inference.
However, enabling end-to-end instance segmentation is not as trivial as adding a mask branch and changing the loss.
We conducted a proof-of-concept experiment by adapting the end-to-end object detection approach to instance segmentation.
The results in Table~\ref{tab1-a} show the inferior performance of doing so.

We argue that the reason behind the failure is the insufficient number of samples for learning the mask head.
On the one hand, the dimensions of masks are much higher than those of classes and boxes.
For example, a mask usually has a $28\times28$ or higher resolution on the COCO dataset, while a bounding box only needs two coordinates to represent.
Therefore, the mask head requires more samples for training.
On the other hand, the proposal bounding boxes obtained by the bipartite matching are usually on a small scale, which also raises the problem of sparse training samples.
For example, Mask R-CNN~\cite{he2017mask} uses $512$ proposal bounding boxes to extract the region of interest (RoI) features for training the mask head, while the number of proposal bounding boxes, \ie, ground truths per image on the COCO dataset, is only $7.7$ on average after bipartite matching.
The gap between the demand and supply of training samples makes the approach prone to failing.

\begin{figure}[t]
\centering
\includegraphics[width=0.98\linewidth]{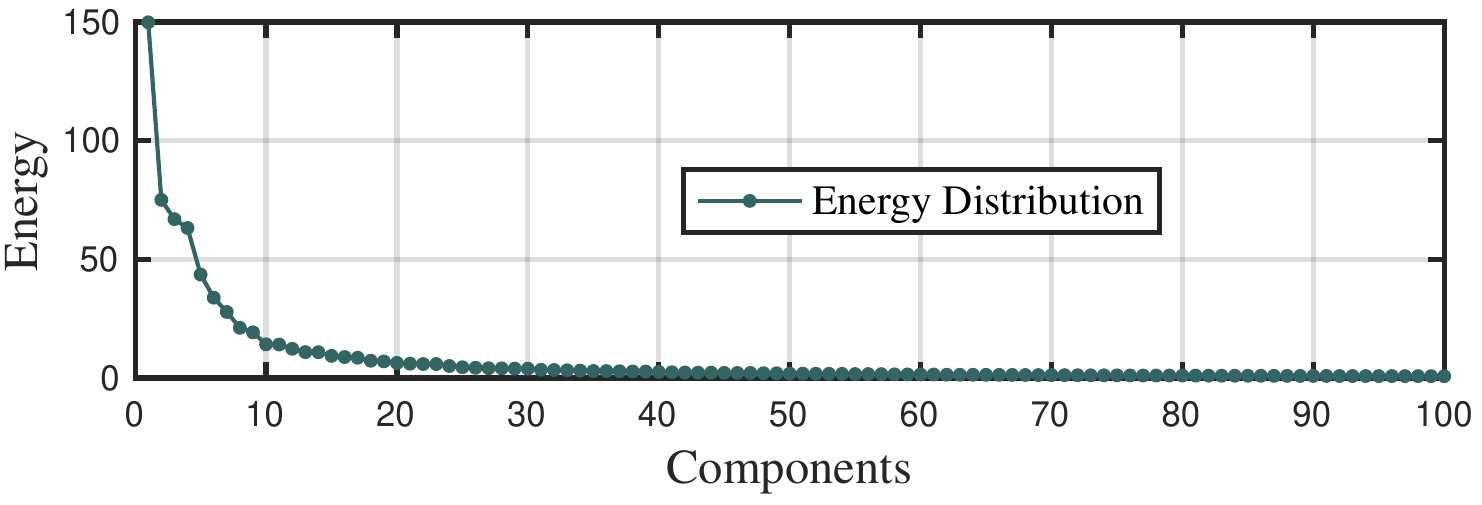}\vspace{-2.mm}
\caption{
\textbf{Component analysis} of masks, ranking the Top@100 components by energy.
The majority of the mask information is embedded into the first few components.
}\vspace{-0.5mm}
\label{fig1}
\end{figure}

While we could blindly augment the ground truth samples to alleviate the problem at the cost of longer training time, we argue that there might be a smarter way.
Not all $28 \times 28$ entries are likely to appear as a mask.
The distribution of the natural masks may lie in a low-dimensional manifold instead of being uniformly scattered.
Based on this intuition, we carried out several dimension reduction experiments on the masks from the training data, and surprisingly found even linear methods, such as principal component analysis (PCA), can do a decent job.
The energy distribution of different components is shown in Fig.~\ref{fig1}.
We observe that the first few components can represent the majority of the mask information.
%
%
Therefore, in this paper, we propose to achieve end-to-end instance segmentation by regressing low-dimensional embeddings instead of raw masks, which enables the training to be effectively conducted with a small number of matched samples.
We also extend the definition of bipartite matching cost based on the mask embeddings.
Furthermore, regressing with the embeddings enables us to design a recurrent refinement strategy that can process detection and segmentation concurrently.
This provides a new way of instance segmentation compared to the top-down and bottom-up frameworks, and boosts the performance.
\begin{figure*}[!ht]
\centering
\includegraphics[width=0.9\linewidth]{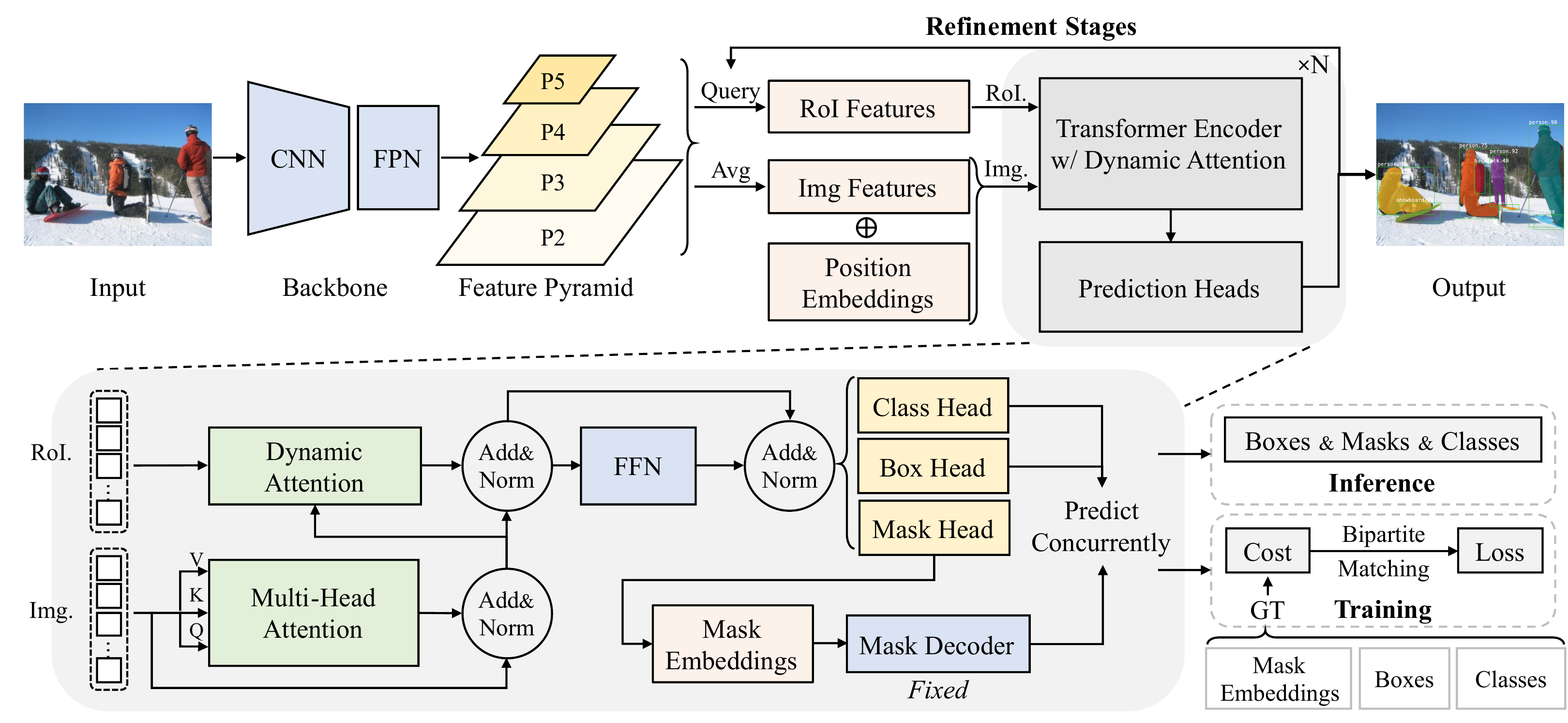}\vspace{-2.mm}
\caption{
\textbf{Framework} of ISTR.
Top: Overview of the pipeline.
Input images are sent to a convolutional neural network (CNN) with a feature pyramid network (FPN)~\cite{lin2017feature} to produce the feature pyramid.
The feature maps from the feature pyramid are cropped and aligned by learnable query boxes with RoIAlign~\cite{he2017mask} to get the RoI features.
Image features are obtained by summing and averaging the feature maps.
Then, a Transformer encoder with dynamic attention fuses the image and RoI features for prediction heads.
The predicted bounding boxes, classes, and masks are recurrently refined in $N$ stages by updating the query boxes.
During training, the predictions are matched with ground truth labels to calculate the set loss.
During inference, the predictions are directly used as the final results without NMS.
Bottom: Details of the modules.
FFN denotes the feed-forward network, and the mask decoder is pre-learned and fixed during training.
}\vspace{-0.5mm}
\label{fig2}
\end{figure*}

Specifically, we propose a new end-to-end instance segmentation framework built upon a Transformer, termed ISTR.
ISTR predicts low-dimensional mask embeddings, and then matches them with ground truth mask embeddings for the set loss.
With the recurrent refinement strategy, ISTR updates the query boxes and refines the set of predictions.
Benefiting from the proposed end-to-end mechanism, we find that even with the suboptimal mask embeddings obtained by the closed-form solution of PCA, ISTR can achieve state-of-the-art performance.
With a single 1080Ti GPU, ISTR obtains a 46.8/38.6 box/mask AP with 13.8 fps using ResNet50-FPN, and a 48.1/39.9 box/mask AP with 11.0 fps using ResNet101-FPN on the \texttt{test-dev} split of the COCO dataset~\cite{lin2014microsoft}.
%
%
Our contributions are summarized as follows:
\begin{itemize}
 \vspace{-5px}
 \item We propose a new framework, termed instance segmentation Transformer (ISTR). For the first time, we demonstrate the potential of using Transformers in end-to-end instance segmentation.\vspace{-5px}
 \item ISTR predicts low-dimensional mask embeddings instead of high-dimensional masks, which facilitates the training with a small number of samples and inspires the design of a bipartite matching cost for masks.\vspace{-5px}
 \item With a recurrent refinement strategy, ISTR concurrently detects and segments instances, providing a new perspective for achieving instance segmentation compared to the bottom-up and top-down frameworks.\vspace{-5px}
 \item Without bells and whistles, ISTR demonstrates accuracy and run-time performance on par with the state-of-the-art methods on the challenging COCO dataset.
\end{itemize}

\section{Related Work}
\paragraph{Instance Segmentation:} Instance segmentation requires instance-level and pixel-level predictions.
Existing works can be summarized into three categories.
Top-down methods~\cite{bolya2019yolact,he2017mask,li2017fully,liu2018path,huang2019mask,chen2019tensormask,chen2020blendmask,lee2020centermask,xie2020polarmask} first detect and then segment the objects.
Bottom-up methods~\cite{yuan2020deep,de2017semantic,liu2017sgn,gao2019ssap} view instance segmentation as a label-then-cluster problem, learning to classify each pixel and then clustering them into groups for each object.
The latest work, SOLO~\cite{wang2020solo,wang2020solov2}, deals with instance segmentation without dependence on box detection.
The proposed ISTR provides a new perspective that directly predicts a set of bounding boxes and mask embeddings, which avoids decomposing instance segmentation into dependent tasks.
Note that the idea of regressing mask embeddings is also investigated in MEInst~\cite{zhang2020mask}.
However, with redundant predictions in each pixel, MEInst obtains suboptimal performance compared to ISTR.
\paragraph{End-to-End Instance-Level Recognition:} Recent studies have revealed the great potential of end-to-end object detection~\cite{hu2018relation,carion2020end,sun2020sparse,wang2020end,zhu2020deformable,zhou2019objects}.
As such, the bipartite matching cost has become an essential component for achieving end-to-end object detection.
For instance segmentation, the works~\cite{ren2017end,romera2016recurrent} explored the end-to-end mechanism with recurrent neural networks.
However, these early trials were only evaluated on small datasets and not against current baselines.
In contrast, ISTR uses the similarity metric of mask embeddings as the bipartite matching cost for masks, and, for the first time, incorporates Transformers~\cite{vaswani2017attention} to improve end-to-end instance segmentation.
\paragraph{Transformers in Computer Vision:} The breakthroughs of Transformers~\cite{vaswani2017attention} in natural language processing have sparked great interest in the computer vision community.
The critical component of Transformer is the multi-head attention, which can significantly enhance the capacity of models~\cite{han2020survey,khan2021transformers}.
So far, Transformers have been successfully used for image recognition~\cite{dosovitskiy2020image,touvron2020training}, object detection~\cite{carion2020end,zhu2020deformable,sun2020sparse}, segmentation~\cite{zheng2020rethinking,ye2019cross}, image super-resolution~\cite{yang2020learning}, video understanding~\cite{sun2019videobert,girdhar2019video}, image generation~\cite{chen2020pre,wang2020sceneformer}, visual question answering~\cite{tan2019lxmert,su2019vl}, and several other tasks~\cite{kumar2021colorization,doersch2020crosstransformers,ye2020few}.
With the sequence information between frames, the contemporary work~\cite{wang2020endvideo} achieves end-to-end video instance segmentation with a Transformer.
Without continuous frames, our study aims to segment instances for a single image, making its design entirely different from the framework of~\cite{wang2020endvideo}.
\paragraph{Multi-Task Learning:} The benefit of learning detection and segmentation jointly was first studied in the work of~\cite{hariharan2014simultaneous}.
After that, Mask R-CNN~\cite{he2017mask} also demonstrated that bounding box detection could benefit from multi-task learning.
Recent works~\cite{chen2018masklab,liu2018path,shen2019cyclic,cao2019triply} have provided more complex mechanisms to improve the performance for multi-tasks.
In our work, we also observe a performance boost when concurrently processing detection and segmentation.
%

\section{Proposed Method}
ISTR aims to directly predict a set of mask embeddings, classes, and bounding boxes for each instance.
To this end, we first introduce a generalized formulation to extract embeddings for representing and reconstructing the masks in Section~\ref{sec3.1}.
A bipartite matching cost and a set loss are introduced in Section~\ref{sec3.2} to pair and regress the predictions with ground truth labels.
Finally, a model that predicts a set of outputs and learns their relations is proposed in Section~\ref{sec3.3}.
The overall framework of ISTR is shown in Fig.~\ref{fig2}.
\subsection{Mask Embeddings}
\label{sec3.1}
To provide a formulation that effectively extracts mask embeddings, we constrain the mutual information between the original and reconstructed masks:
\begin{equation}
\begin{split}
\label{eq1}
\max \mathcal{I}\Big(\boldsymbol{M}, f\big(g(\boldsymbol{M})\big)\Big),
\end{split}
\end{equation}
where $\mathcal{I}(\cdot, \cdot)$ denotes the mutual information between two random variables, $\boldsymbol{M}$ denotes a set of masks $\{\boldsymbol{m}_i\in\mathbb{R}^{s^2}|i=1,...,n\}$, $s^2$ is the dimension of masks, $g(\cdot)$ denotes the mask encoder for extracting embeddings and $f(\cdot)$ denotes the mask decoder for reconstructing masks.
Eq.~\ref{eq1} guarantees that the encoding and decoding phases have minimal information loss, which implicitly encourages the embeddings to represent the masks.
After derivation, we have a generalized objective function for the mask embeddings:
\begin{equation}
\begin{split}
\label{eq2}
\min\sum_{i=1}^n||\boldsymbol{m}_i - f(\boldsymbol{r}_i)||_2^2,
\end{split}
\end{equation}
where $\boldsymbol{r}_i=g\big(\boldsymbol{m}_i\big)$ denotes the mask embeddings, and $||\cdot||_2$ is the L2-norm.
By making the functions of the encoder and decoder simple linear transformations via a matrix $\boldsymbol{D}\in\mathbb{R}^{s^2\times l}$, \ie, $f\big(g(\boldsymbol{m}_i)\big)=\boldsymbol{D}\boldsymbol{D}^T\boldsymbol{m}_i$ and $\boldsymbol{DD}^T=\boldsymbol{I}_l$, the objective function becomes:
\begin{equation}
\begin{split}
\label{eq3}
\boldsymbol{D}^*=\arg\min_{\boldsymbol{D}}\sum_{i=1}^n||\boldsymbol{m}_i - \boldsymbol{DD}^T\boldsymbol{m}_i||_2^2,
\end{split}
\end{equation}
where $l$ is the dimension of the mask embeddings, and $\boldsymbol{I}_l$ denotes the $l\times l$ unit matrix.
Eq.~\ref{eq3} has the same formulation as the objective function of PCA, which provides a closed-form solution to learn the transformation.
Note that the objective function in Eq.~\ref{eq2} can also be optimized by other models, such as an autoencoder.
\subsection{Matching Cost and Prediction Loss}
\label{sec3.2}
After obtaining the encoder and decoder for mask embeddings, we define a bipartite matching cost and a set prediction loss for end-to-end instance segmentation.
Let us denote the ground truth bounding boxes, classes, and masks as $\boldsymbol{Y}=\{\boldsymbol{b}_i,\boldsymbol{c}_i,\boldsymbol{m}_i|i=1,...,n\}$.
The predicted bounding boxes, classes, and mask embeddings are denoted as $\widetilde{\boldsymbol{Y}}=\{\widetilde{\boldsymbol{b}}_i,\widetilde{\boldsymbol{c}}_i,\widetilde{\boldsymbol{r}}_i|i=1,...,k\}$, where $k>n$.
\paragraph{Bipartite Matching Cost:} For the bipartite matching, we search for a permutation of $n$ non-repeating integers $\boldsymbol{\sigma}\in\{1,2,...,k\}$ with the lowest cost, as:
\begin{equation}
\begin{split}
\label{eq4}
\boldsymbol{\sigma}^*=\arg\min_{\boldsymbol{\sigma}}\sum_{i=1}^n\big(&\mathcal{C}_{box}(\boldsymbol{b}_i, \widetilde{\boldsymbol{b}}_{\sigma(i)})\\
+\mathcal{C}_{cls}(&\boldsymbol{c}_i,\widetilde{\boldsymbol{c}}_{\sigma(i)})+\mathcal{C}_{mask}(\boldsymbol{m}_i,\widetilde{\boldsymbol{r}}_{\sigma(i)})\big).
\end{split}
\end{equation}
Inspired by~\cite{carion2020end,sun2020sparse}, we define the matching cost for bounding boxes as:
\begin{equation}
\begin{split}
\label{eq5}
\mathcal{C}_{box}(\boldsymbol{b}_i, \widetilde{\boldsymbol{b}}_{\sigma(i)})&=\lambda_{L1}\cdot\mathcal{C}_{L1}(\boldsymbol{b}_i, \widetilde{\boldsymbol{b}}_{\sigma(i)})\\
&+\lambda_{giou}\cdot\mathcal{C}_{giou}(\boldsymbol{b}_i, \widetilde{\boldsymbol{b}}_{\sigma(i)}),
\end{split}
\end{equation}
and the matching cost for classes as:
\begin{equation}
\begin{split}
\label{eq6}
\mathcal{C}_{cls}(\boldsymbol{c}_i, \widetilde{\boldsymbol{c}}_{\sigma(i)})&=-\lambda_{cls}\cdot \widetilde{p}_{\sigma(i)}(\boldsymbol{c}_i),
\end{split}
\end{equation}
where $\lambda$ denotes the hyperparameters that balance the costs, $\mathcal{C}_{L1}(\cdot,\cdot)$ denotes the L1 cost, $\mathcal{C}_{giou}(\cdot,\cdot)$ denotes the generalized IoU~\cite{rezatofighi2019generalized} cost, and $\widetilde{p}_{\sigma(i)}(\boldsymbol{c}_i)$ is the probability of class $\boldsymbol{c}_i$.
Instead of directly matching the high-dimensional masks, we use the similarity metric between mask embeddings to match them, which is defined as:
\begin{equation}
\begin{split}
\label{eq7}
\mathcal{C}_{mask}(\boldsymbol{m}_i, \widetilde{\boldsymbol{r}}_{\sigma(i)})&=\\
-\frac{1}{2}\lambda_{mask}\cdot &\big(<\frac{\widetilde{\boldsymbol{r}}_{\sigma(i)}}{||\widetilde{\boldsymbol{r}}_{\sigma(i)}||_2},\frac{g(\boldsymbol{m}_i)}{||g(\boldsymbol{m}_i)||_2}>+1\big),
\end{split}
\end{equation}
where the mask embeddings are L2 normalized, and the dot product between two normalized vectors is used to calculate the cosine similarity.
We add $1$ to the result and divide it by $2$ to guarantee that the values are in the range of $[0,1]$.
\paragraph{Set Prediction Loss:} For the set prediction loss, we use the matched predictions to regress the ground truth targets.
The set prediction loss is defined as:
\begin{equation}
\begin{split}
\label{eq8}
\mathcal{L}_{set}(\boldsymbol{Y}, \widetilde{\boldsymbol{Y}},\boldsymbol{\sigma}^*)&=\frac{1}{n}\sum_{i=1}^n\big(\mathcal{L}_{box}(\boldsymbol{b}_i, \widetilde{\boldsymbol{b}}_{\sigma^*(i)})\\
+\mathcal{L}_{cls}(&\boldsymbol{c}_i,\widetilde{\boldsymbol{c}}_{\sigma^*(i)})
+\mathcal{L}_{mask}(\boldsymbol{m}_i,\widetilde{\boldsymbol{r}}_{\sigma^*(i)})\big),
\end{split}
\end{equation}
where $\mathcal{L}_{box}(\cdot,\cdot)$ is defined the same as $\mathcal{C}_{box}(\cdot,\cdot)$, and $\mathcal{L}_{cls}(\cdot,\cdot)$ is the focal loss~\cite{lin2017focal} for classification.
For masks, we add the dice loss~\cite{milletari2016v} to improve the learned embeddings for reconstructing the masks.
The mask loss is defined as:
\begin{equation}
\begin{split}
\label{eq9}
\mathcal{L}_{mask}(\boldsymbol{m}_i,\widetilde{\boldsymbol{r}}_{\sigma^*(i)})=&\lambda_{mask}\cdot\Big(\mathcal{L}_{L2}(g(\boldsymbol{m}_i),\widetilde{\boldsymbol{r}}_{\sigma^*(i)})\\
&+\mathcal{L}_{dice}\big(\boldsymbol{m}_i,f(\widetilde{\boldsymbol{r}}_{\sigma^*(i)})\big)\Big),
\end{split}
\end{equation}
where $\mathcal{L}_{L2}(\cdot,\cdot)$ denotes the L2 loss and $\mathcal{L}_{dice}(\cdot,\cdot)$ denotes the dice loss.
\begin{algorithm}[t]
\caption{Instance Segmentation Transformer}
\label{alg1}
\begin{algorithmic}
 \STATE \textit{//\ ---------\ Training Phase\ ---------}
 \STATE {\bfseries Input:} Images and ground truth labels.
 \STATE {\bfseries Output:} Learned ISTR model.

 \STATE \textit{1:} Learn the mask encoder and decoder via Eq.~\ref{eq2}.
 \STATE \textit{2:} Initialize the learnable query boxes $\widetilde{\boldsymbol{B}}^0$.
 \STATE \textit{3:} \textbf{Repeat}
 \STATE \textit{4:}\ \ \ \ \textbf{For} $i=1,2,..,N$ stage:
 \STATE \textit{5:}\ \ \ \ \ \ \ \ Obtain RoI features by RoIAlign with $\widetilde{\boldsymbol{B}}^{i-1}$.
 \STATE \textit{6:}\ \ \ \ \ \ \ \ Predict via ISTR encoder and heads.
 \STATE \textit{7:}\ \ \ \ \ \ \ \ Match predictions with labels via Eq.~\ref{eq4}.
 \STATE \textit{8:}\ \ \ \ \ \ \ \ Calculate loss via Eq.~\ref{eq8} and train ISTR.
 \STATE \textit{9:}\ \ \ \ \ \ \ \ Update $\widetilde{\boldsymbol{B}}^i$ from $\widetilde{\boldsymbol{B}}^{i-1}$ with the predicted boxes.
 \STATE \textit{10:} \textbf{Until} scheduled epochs.
 \STATE \textit{//\ ---------\ Inference Phase\ ---------}
 \STATE {\bfseries Input:} Images to be processed.
 \STATE {\bfseries Output:} Detected and segmented objects.
 \STATE \textit{1:} \textbf{For} $i=1,2,..,N$ stage:
 \STATE \textit{2:}\ \ \ \ Obtain RoI features by RoIAlign with $\widetilde{\boldsymbol{B}}^{i-1}$.
 \STATE \textit{3:}\ \ \ \ Obtain predictions via ISTR encoder and heads.
 \STATE \textit{4:}\ \ \ \ Update $\widetilde{\boldsymbol{B}}^i$ from $\widetilde{\boldsymbol{B}}^{i-1}$ with the predicted boxes.
 \STATE \textit{5:} Output the set of predictions in the final stage.
 \end{algorithmic}
\end{algorithm}
\begin{table*}[t]
\center
\subfloat[\textbf{Mask \vs Mask Embeddings:} Regression with mask embeddings instead of masks brings better performance to the mask APs. The performance improves when the embedding dimension $l$=60, and saturates when the dimension $l$=80. Directly expanding the mask as embeddings, \ie, $l$=784, has worse performance.\label{tab1-a}]{
\tablestyle{4.1pt}{1.2}\begin{tabular}{c|x{22}x{22}x{22}|x{22}x{22}x{22}}
& AP$^{m}$ & AP$^{m}_{50}$ & AP$^{m}_{75}$ & AP$^{b}$ & AP$^{b}_{50}$ & AP$^{b}_{75}$ \\ \shline
\textit{mask} & 31.6 & 53.3 & 33.0 & 40.2 & 58.4 & 43.6 \\ \hline
$l$=40 & 33.7 & \textbf{55.6} & 35.5 & \textbf{41.1} & 58.9 & \textbf{44.8} \\
$l$=60 & \textbf{34.2} & \textbf{55.6} & \textbf{36.4} & 41.0 & 58.9 & 44.4 \\
$l$=80 & 33.8 & 55.3 & 35.9 & 40.6 & 58.6 & 44.1 \\
$l$=784 & 31.4 & 55.2 & 31.8 & \textbf{41.1} & \textbf{60.0} & 44.4
\end{tabular}}\hspace{4mm}\vspace{-1.5mm}
\subfloat[\textbf{Mask Cost Functions:} Matching with the dice loss between the predicted and ground truth masks performs slightly better than w/o the mask cost. The L1 loss between the predicted and encoded mask embeddings also has a slight improvement. Using cosine similarity as the mask cost function brings expected gains.\label{tab1-b}]{
\tablestyle{4.88pt}{1.2}\begin{tabular}{c|x{22}x{22}x{22}|x{22}x{22}x{22}}
& AP$^{m}$ & AP$^{m}_{50}$ & AP$^{m}_{75}$ & AP$^{b}$ & AP$^{b}_{50}$ & AP$^{b}_{75}$ \\ \shline
w/o & 33.8 & 55.5 & 35.7 & 40.7 & 58.8 & 44.1 \\ \hline
dice & 33.9 & 55.4 & 35.8 & 40.8 & 58.8 & 44.3 \\
L1 & 34.0 & 55.5 & 35.6 & 40.9 & \textbf{58.9} & 44.3 \\
cosine & \textbf{34.2} & \textbf{55.6} & \textbf{36.4} & \textbf{41.0} & \textbf{58.9} & \textbf{44.4} \\
\multicolumn{7}{l}{} 
\end{tabular}} \\%
\subfloat[\textbf{Loss Functions:} Learning masks with both a pixel-level dice loss and a embedding-level L2 loss yields gains in mask APs.\label{tab1-c}]{
\tablestyle{3.6pt}{1.2}\begin{tabular}{c|x{22}x{22}x{22}|x{22}x{22}x{22}}
& AP$^{m}$ & AP$^{m}_{50}$ & AP$^{m}_{75}$ & AP$^{m}_S$ & AP$^{m}_{M}$ & AP$^{m}_{L}$ \\ \shline
dice & 32.6 & 55.0 & 33.5 & 17.3 & 34.8 & 47.6 \\
L2 & 33.8 & 55.5 & 35.5 & 17.3 & 36.1 & 49.5 \\ \hline
L2+dice & \textbf{34.2} & \textbf{55.6} & \textbf{36.4} & \textbf{17.6} & \textbf{36.5} & \textbf{50.6}
\end{tabular}}\hspace{4mm}\vspace{-1.5mm}
\subfloat[\textbf{Attention Type:} Dynamic attention brings significant gains compared with multi-head attention in fusing the RoI and image features.\label{tab1-d}]{
\tablestyle{3.8pt}{1.2}\begin{tabular}{c|x{22}x{22}x{22}|x{22}x{22}x{22}}
& AP$^{m}$ & AP$^{m}_{50}$ & AP$^{m}_{75}$ & AP$^{b}$ & AP$^{b}_{50}$ & AP$^{b}_{75}$ \\ \shline
\multicolumn{1}{c|}{multi-head} & 22.0 & 43.9 & 19.6 & 31.2 & 50.5 & 32.5 \\ 
dynamic & \textbf{34.2} & \textbf{55.6} & \textbf{36.4} & \textbf{41.0} & \textbf{58.9} & \textbf{44.4} \\ \hline
& \textit{+12.2} & \textit{+11.7} & \textit{+16.8} & \textit{+9.8} & \textit{+8.4} & \textit{+11.9}
\end{tabular}} \\
\subfloat[\textbf{Pooling Type:} The global average pooling yields better performance than the global max-pooling. Combining the image features with position embeddings increases the performance in both mask and box APs.\label{tab1-e}]{
\tablestyle{3.65pt}{1.2}\begin{tabular}{c|x{12}x{12}|x{12}|x{22}x{22}x{22}|x{22}x{22}x{22}|x{22}x{22}x{22}|x{22}x{22}x{22}}
\multicolumn{1}{c|}{} & img. & type & pos. & AP$^{m}$ & AP$^{m}_{50}$ & AP$^{m}_{75}$ & AP$^{m}_S$ & AP$^{m}_M$ & AP$^{m}_L$ & AP$^{b}$ & AP$^{b}_{50}$ & AP$^{b}_{75}$ & AP$^{b}_S$ & AP$^{b}_M$ & AP$^{b}_L$ \\ \shline
\multirow{4}{*}{\emph{features}} & \checkmark & max & & 33.3 & 54.6 & 35.2 & 17.4 & 35.6 & 48.5 & 40.0 & 57.6 & 43.5 & 24.2 & 42.6 & 52.6 \\
& \checkmark & avg & & 33.6 & 55.3 & 35.5 & 17.3 & 36.4 & 49.5 & 40.6 & 58.8 & 44.1 & 23.9 & 43.2 & 54.1 \\
& - & - & \checkmark & 34.0 & 55.3 & 36.3 & 17.4 & 36.4 & 50.5 & 40.9 & 58.8 & 44.3 & 24.3 & 43.2 & 55.0 \\
& \checkmark & avg & \checkmark & \textbf{34.2} & \textbf{55.6} & \textbf{36.4} & \textbf{17.6} & \textbf{36.5} & \textbf{50.6} & \textbf{41.0} & \textbf{58.9} & \textbf{44.4} & \textbf{24.8} & \textbf{43.3} & \textbf{55.3} 
\end{tabular}}\vspace{-2.mm}
\caption{\textbf{Ablations.} We train on the COCO \texttt{train2017} split, and report \emph{mask} as well as \emph{box} APs on the \texttt{val2017} split.}\vspace{-0.5mm}
\label{tab1}
\end{table*}
\subsection{Instance Segmentation Transformer}
\label{sec3.3}
The architecture of ISTR is depicted in Fig.~\ref{fig2}. It contains four main components: a CNN backbone with FPN~\cite{lin2017feature} to extract features for each instance, a Transformer encoder with dynamic attention to learn the relations between objects, a set of prediction heads that conduct detection and segmentation concurrently, as well as the $N$-step recurrent update for refining the set of predictions.
\paragraph{Backbone:} We use a CNN backbone with FPN to extract the features ranging from P2 to P5 level of the feature pyramid.
Then, $k$ learnable query boxes $\widetilde{\boldsymbol{B}}^0=\{\widetilde{\boldsymbol{b}}_i^0|i=1,...,k\}$ initially covering the whole images are used to extract $k$ RoI features $\boldsymbol{U}^0\in\mathbb{R}^{k\times d\times t\times t}$ via RoIAlign~\cite{he2017mask}.
Image features $\boldsymbol{P}\in\mathbb{R}^{k\times d}$ are first extracted by averaging and summing the features from P2 to P5, and then expand the first dimension to $k$ for each RoI feature.
Learnable position embeddings $\boldsymbol{E}\in\mathbb{R}^{k\times d}$ are initialized randomly.
%
%
\paragraph{Transformer Encoder and Dynamic Attention:} The sum of image features and position embeddings is first transformed by three learnable weight matrices to obtain the inputs $\boldsymbol{Q}=(\boldsymbol{P}+\boldsymbol{E})\boldsymbol{W}_Q,\boldsymbol{K}=(\boldsymbol{P}+\boldsymbol{E})\boldsymbol{W}_K,\boldsymbol{V}=(\boldsymbol{P}+\boldsymbol{E})\boldsymbol{W}_V$ for the self-attention module defined as:
\begin{equation}
\begin{split}
\label{eq10}
\boldsymbol{Z}=softmax(\frac{\boldsymbol{QK}^T}{\sqrt{d}})\boldsymbol{V}.
\end{split}
\end{equation}
The multi-head attention comprises multiple self-attention blocks, \eg, eight in the original Transformer~\cite{vaswani2017attention}, to encapsulate multiple complex relationships amongst different features.
Inspired by~\cite{sun2020sparse}, we add a dynamic attention module for better fusing the RoI and image features, which is defined as the attention conditioned on the RoI features $\boldsymbol{U}^i$ in the $i$-th step:
\begin{equation}
\begin{split}
\label{eq11}
\boldsymbol{O}^i=\boldsymbol{U}^i\cdot fc(\boldsymbol{Z}),
\end{split}
\end{equation}
where $fc(\cdot)$ denotes a fully connected layer to generate the dynamic parameters.
The obtained features $\boldsymbol{O}^{i}$ are then used in the prediction heads to produce the set of outputs.
\paragraph{Prediction Heads:} The set of predictions is computed by the heads, including a class head, a box head, a mask head, and a fixed mask decoder.
The box head predicts the residual values of normalized center coordinates, height, and width for updating the query boxes $\widetilde{\boldsymbol{B}}^i$ in the $i$-th step, and the class head predicts the classes using a softmax function.
The mask head outputs the mask embeddings, which are then reconstructed to predict masks using the pre-learned mask decoder.
\begin{figure*}[t]
\centering
\subfloat[\textbf{Qualitative Results:} Masks and bounding boxes are from ISTR using ResNet101-FPN on the COCO \texttt{val2017} split, with a threshold of $0.4$.\label{fig3-3}]{
\includegraphics[width=0.97\linewidth]{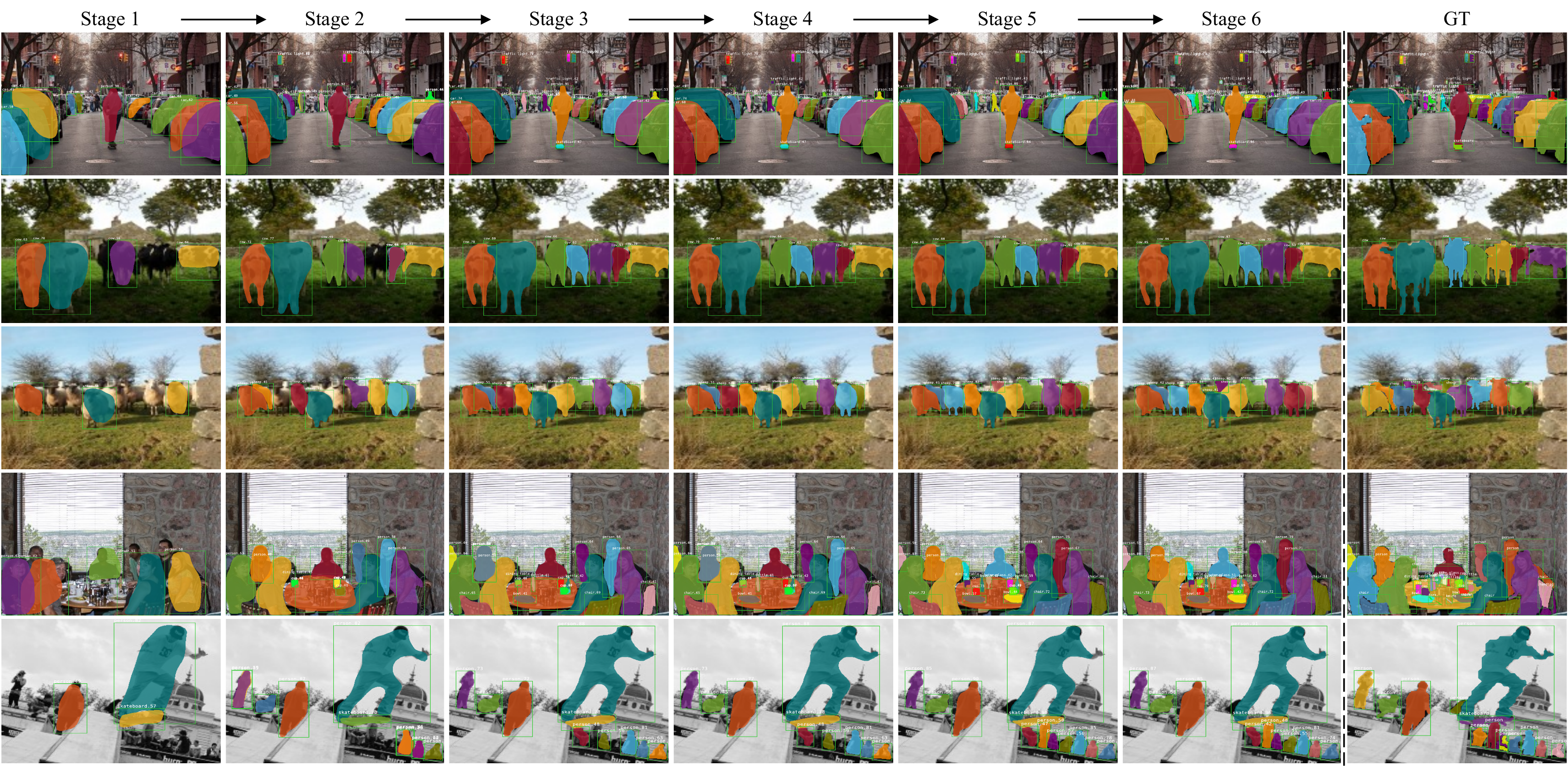}}\vspace{-3.mm}\\
\subfloat[AP results on the COCO \texttt{val2017} split with \textbf{ResNet50-FPN}.\label{fig3-1}]{
\includegraphics[width=0.47\linewidth]{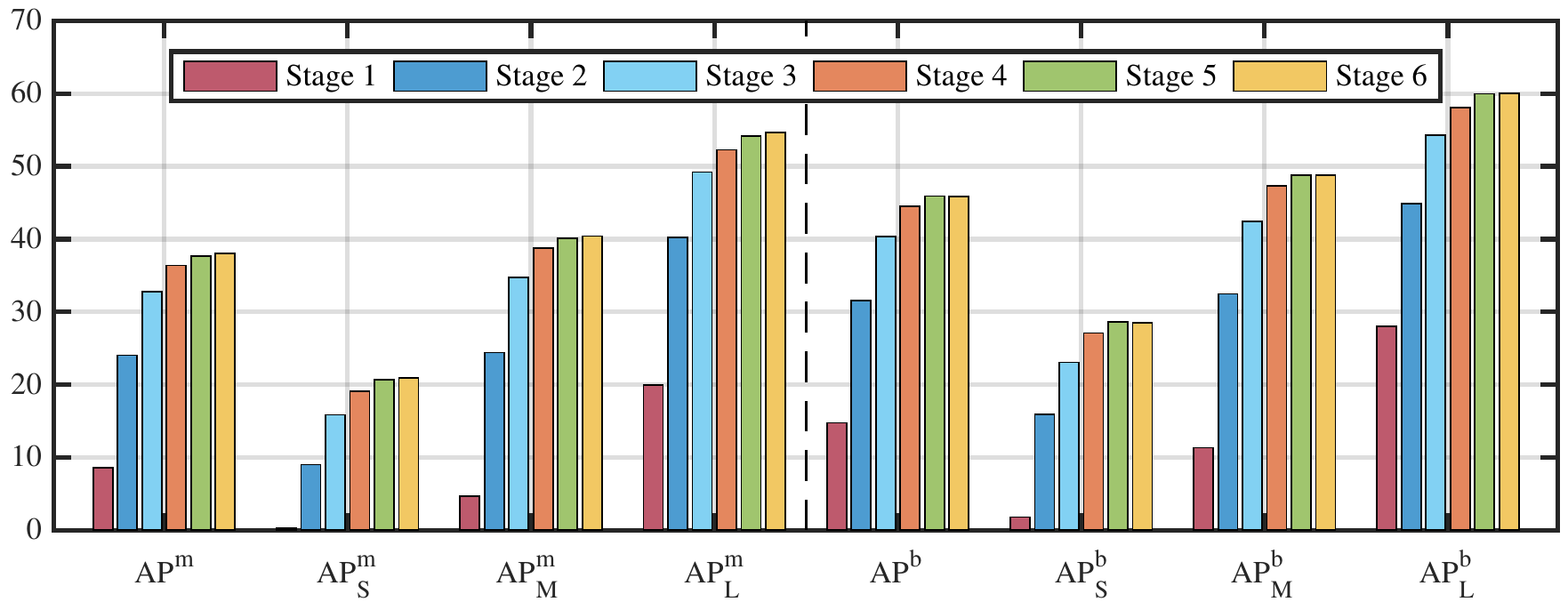}}\hspace{4mm}
\subfloat[AP results on the COCO \texttt{val2017} split with \textbf{ResNet101-FPN}.\label{fig3-2}]{
\includegraphics[width=0.47\linewidth]{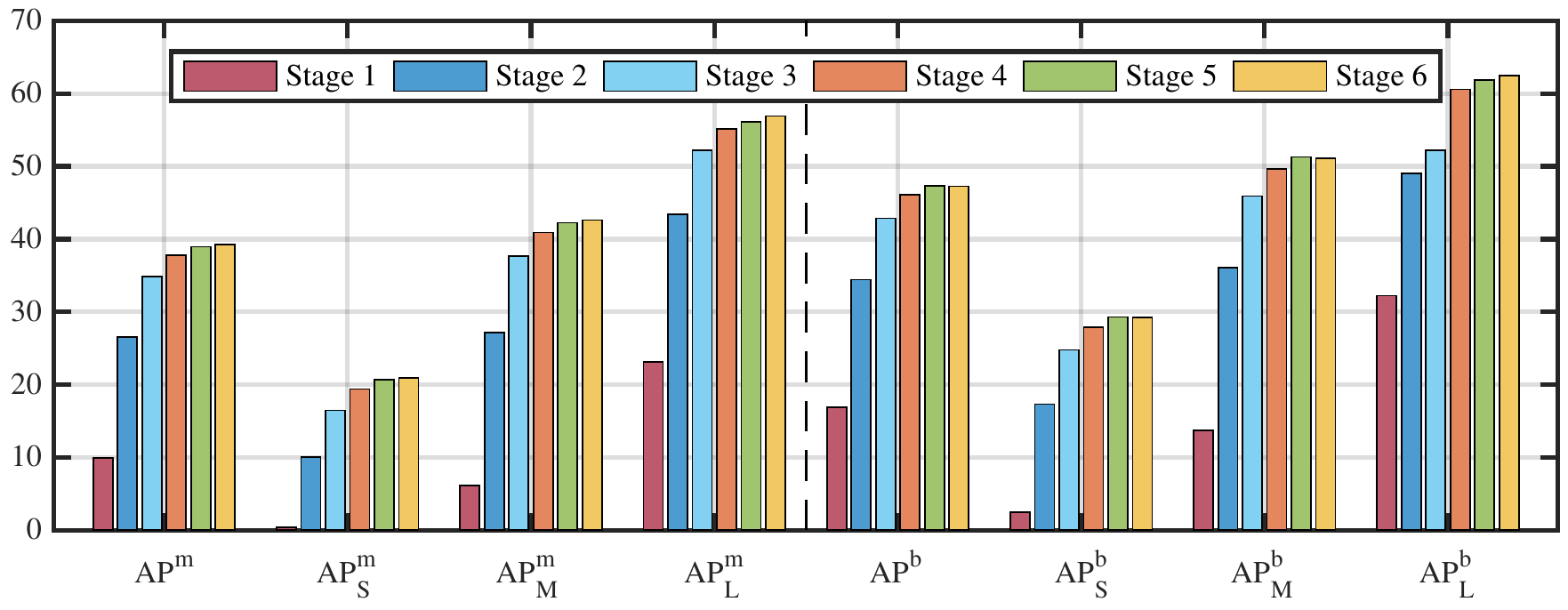}}\vspace{-2.mm}
\caption{
\textbf{Stage Analysis.} We report quantitative and qualitative results of ISTR for $N=6$ stages. Both results show that the bounding boxes and masks are refined stage by stage, and the performance saturates in the last two stages.
}\vspace{-0.5mm}
\label{fig3}
\end{figure*}
\paragraph{Recurrent Refinement Strategy:} The query boxes $\widetilde{\boldsymbol{B}}^i$ are recurrently updated by the predicted boxes, which refines the predictions and makes it possible to process the detection and segmentation concurrently.
The overall process is summarized in Alg.~\ref{alg1}.
\section{Experiments}
\paragraph{Dataset and Evaluation Metrics:} Our experiments are performed on the MS COCO dataset~\cite{lin2014microsoft}, which contains 123K images with 80-class instance labels.
Our models are trained on the \texttt{train2017} split (118K images), and the ablation study is carried out on the \texttt{val2017} split (5K images).
Final results are reported on the \texttt{test-dev} split, which has no public labels and is evaluated on the server.
We report the standard COCO metrics including AP (\ie, averaged over IoU thresholds), AP$_{50}$, AP$_{75}$, and AP$_S$, AP$_M$, AP$_L$ (\ie, AP at different scales) for both boxes and masks, denoted as AP$^b$ and AP$^m$, respectively.
\begin{table*}[t]
\center
\tablestyle{3.8pt}{1.2}\begin{tabular}{l|c|c|x{22}x{22}x{22}x{22}|x{22}x{22}x{22}x{22}|c|c|c}
method & backbone & Epochs & AP$^{m}$ & AP$^{m}_{S}$ & AP$^{m}_{M}$ & AP$^{m}_{L}$ & AP$^{b}$ & AP$^{b}_{S}$ & AP$^{b}_{M}$ & AP$^{b}_{L}$ & FPS & Time & GPU\\ \shline
Mask R-CNN~\cite{he2017mask} & ResNet50-FPN & 36& 37.5 & 21.1 & 39.6 & 48.3 & 41.3 & 24.2 & 43.6 & 51.7 & 15.3 & 65.6 & 1080Ti \\
MEInst~\cite{zhang2020mask} & ResNet50-FPN &36 & 33.5 & 19.3 & 35.7 & 42.1 & 42.5 & 25.6 & 45.1 & 52.2 & 15.0 & 66.8 & 1080Ti\\
CondInst~\cite{tian2020conditional} & ResNet50-FPN & 36 & 37.8 & 21.0 & 40.3 & 48.7 & 42.1 & 25.1 & 44.5 & 52.1 & 15.4 & 65.0 & 1080Ti\\
BlendMask~\cite{chen2020blendmask} & ResNet50-FPN & 36 & 37.8 & 18.8 & 40.9 & 53.6 & 43.0 & 25.3 & 45.4 & 54.0 & 15.0 & 66.8 & 1080Ti\\
SOLOv2~~\cite{wang2020solov2} & ResNet50-FPN & 36 & 38.2 & 16 & \textbf{41.2} & \textbf{55.4} & 40.7 & 18.4 & 43.5 & 57.6 & 10.5 & 95.5 & 1080Ti\\ \hline
DETR~~\cite{carion2020end} & ResNet50 & 500 & - & - & - & - & 42.0 & 20.5 & 45.8 & \textbf{61.1} & - & - & - \\
Sparse R-CNN~~\cite{sun2020sparse} & ResNet50-FPN & 36 & - & - & - & - & 44.5 & 26.9 & 47.2 & 59.5 & - & - & - \\\hline
\textbf{ISTR}, \emph{ours} & ResNet50-FPN & 36 & \textbf{38.6} & \textbf{22.1} & 40.4 & 50.6 & \textbf{46.8} & \textbf{27.8} & \textbf{48.7} & 59.9 & 13.8 & 72.5 & 1080Ti\\ \hline\hline
Mask R-CNN~\cite{he2017mask} & ResNet101-FPN & 36 & 38.8 & 21.8 & 41.4 & 50.5 & 43.1 & 25.1 & 46.0 & 54.3 & 11.8 & 85.0 & 1080Ti\\
MEInst~\cite{zhang2020mask} & ResNet101-FPN & 36 & 35.3 & 20.4 & 37.8 & 44.5 & 44.5 & 26.8 & 47.3 & 54.9 & 11.2 & 89.3 & 1080Ti\\
CondInst~\cite{tian2020conditional} & ResNet101-FPN & 36 & 39.1 & 21.5 & 41.7 & 50.9 & 43.5 & 25.8 & 46.0 & 54.1 & 12.0 & 83.2 & 1080Ti\\
BlendMask~\cite{chen2020blendmask} & ResNet101-FPN & 36 & 39.6 & 22.4 & 42.2 & 51.4 & 44.7 & 26.6 & 47.5 & 55.6 & 11.5 & 86.6 & 1080Ti\\
SOLOv2~~\cite{wang2020solov2} & ResNet101-FPN & 36 & 39.7 & 17.3 & \textbf{42.9} & \textbf{57.4} & 42.6 & 22.3 & 46.7 & 56.3 & 9.0 & 111.6 & 1080Ti\\ \hline
DETR~~\cite{carion2020end} & ResNet101 & 500 & - & - & - & - & 43.5 & 21.9 & 48.0 & \textbf{61.8} & - & - & - \\
Sparse R-CNN~~\cite{sun2020sparse} & ResNet101-FPN & 36 & - & - & - & - & 45.6 & \textbf{28.3} & 48.3 & 61.6 & - & - & - \\\hline
\textbf{ISTR}, \emph{ours} & ResNet101-FPN & 36 & \textbf{39.9} & \textbf{22.8} & 41.9 & 52.3 & \textbf{48.1} & \textbf{28.7} & \textbf{50.4} & 61.5 & 11.0 & 91.3 & 1080Ti
\end{tabular}\vspace{-2.mm}
\caption{\textbf{Quantitative Results} of ISTR on the COCO \texttt{test-dev} split. All the models are learned by multi-scale training. The results of FPS are measured with a single 1080Ti GPU. The performance of Mask R-CNN is the result of the modified version with implementation details in TensorMask~\cite{chen2019tensormask}.}\vspace{-0.5mm}
\label{tab2}
\end{table*}
\paragraph{Training Details:} We follow the strategy in~\cite{zhang2020mask,tian2019decoders} to encode the ground truth mask embeddings.
ResNet50 and ResNet101~\cite{he2016deep} pre-trained on ImageNet~\cite{deng2009imagenet} are used as our backbone networks.
FPN~\cite{lin2017feature} is used to extract the feature pyramid.
For the ablation study, all the models are trained over $12$ epochs with learning rate decay, dividing by $10$ at epoch $9$ and $11$, respectively.
All results in the ablation study are tested with ResNet50-FPN and reported on the COCO \texttt{val2017} split.
The training schedule for the final models is $36$ epochs with the learning rate divided by $10$ at epoch $27$ and $33$, respectively.
The mini-batch contains $16$ images, and the models are trained with eight GPUs.
Following~\cite{sun2020sparse}, we use the AdamW~\cite{loshchilov2017decoupled} optimizer and an initial learning rate of $0.000025$.
The input images are resized such that the shortest side is at least $480$ and at most $800$ pixels, while the longest side is at most $1333$.
The number of predictions, \ie, $k$, is set to $300$, and the number of self-attention blocks in the multi-head attention is set to $8$.
The number of recurrent refinement stages is set to $6$.
Following~\cite{carion2020end,zhu2020deformable}, we set $\lambda_{cls}$, $\lambda_{L1}$, and $\lambda_{giou}$ to $2$, $5$, and $2$, respectively.
The hyperparameter $\lambda_{mask}$ is set to $2$.
%
\subsection{Ablation Experiments}
To analyze ISTR, we conduct ablation studies on the choices of mask embeddings, cost functions, loss functions, the effect of position embeddings, and pooling types.
Results are shown in Table~\ref{tab1} and discussed in detail next.
\paragraph{Mask \vs Mask Embeddings:} Table~\ref{tab1-a} shows the models with various types of mask representations, including the original masks with $28\times 28$ dimensions and the mask embeddings with different dimensions $l$.
We also expand the original masks to vectors to test the performance.
Using raw masks for segmentation is implemented with the mask head from Mask R-CNN in a top-down manner, and other settings are the same as ISTR.
We obtain the following results.
First, predicting mask embeddings brings better performance to the mask APs than predicting masks.
The results verify our concern that the high-dimensional masks cannot be effectively learned with a small number of matched samples.
In contrast, the mask embeddings can be well regressed, as their dimensions are much lower than those of masks, \eg, $40$, $60$, $80$ \vs $784$.
Second, the performance of mask embeddings improves when the dimension $l$=60 and saturates when the dimension $l$=80.
Finally, regressing with the expanded masks, \ie, $l$=784, as embeddings has worse performance in mask APs.
%
\paragraph{Cost Functions:} Appropriate cost functions can match high-quality predictions with ground truth labels for the set loss.
In Table~\ref{tab1-b}, we compare the performance with different choices of cost functions for the masks.
Matching the predicted masks with ground truth masks by the dice loss does not produce expected gains compared to matching without the mask cost function.
Using the L1 loss between embeddings slightly improves performance, and using the cosine similarity between embeddings as the mask cost function brings expected gains.
%
%
\paragraph{Loss Functions:} We investigate two types of loss functions for ISTR: the dice loss at the pixel-level and the L2 loss at the embedding-level.
The dice loss is calculated using the masks reconstructed by the mask decoder.
As shown in Table~\ref{tab1-c}, only calculating the dice loss at the pixel-level without constraining the mask embeddings has an inferior performance.
Although the L2 loss between the predicted and encoded mask embeddings improves the performance, the learned mask embeddings are slightly suboptimal for reconstructing the original masks.
Therefore, training with both the pixel-level dice loss and the embedding-level L2 loss produces better results.
\paragraph{Attention Type:} We next study the effect of the dynamic attention module, which is used to fuse the RoI and image features, by replacing it with the multi-head attention module.
As shown in Table~\ref{tab1-d}, the dynamic attention module performs much better than the multi-head attention module.
We believe this may be because the multi-projection in the multi-head attention complicates the fusion of RoI and image features, which is essential for learning the relations between objects.
From the results, we infer that a single projection is more effective for learning such relations.
\begin{figure*}[t]
\centering
\subfloat[Mask R-CNN~\cite{he2017mask} (top) \vs ISTR (bottom) using ResNet101-FPN. Mask R-CNN suffers inferior segmentation when bad duplicate removal occurs.\label{fig4-1}]{
\includegraphics[width=0.97\linewidth]{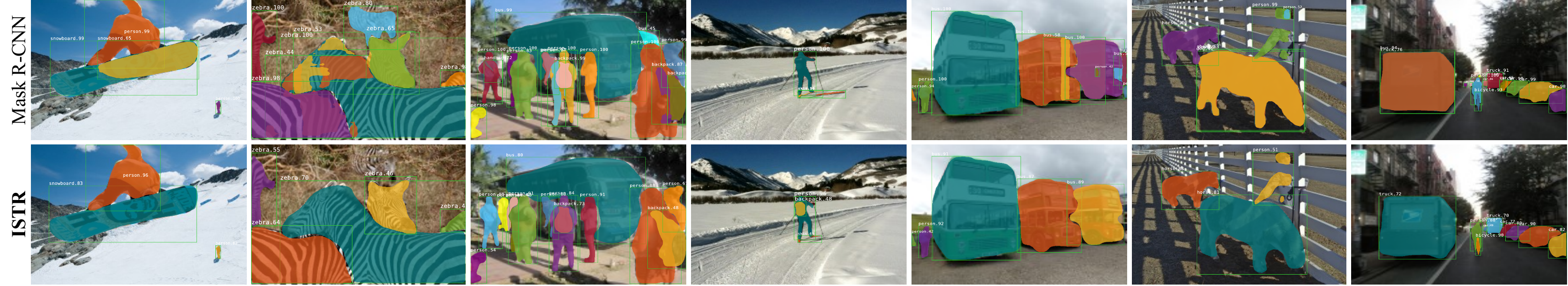}}\vspace{-3mm}\\
\subfloat[More visualization results of \textbf{ISTR}, using ResNet101-FPN and running at 11.0 fps on a 1080Ti GPU, with 39.9 mask AP (Table~\ref{tab2}).\label{fig4-2}]{
\includegraphics[width=0.97\linewidth]{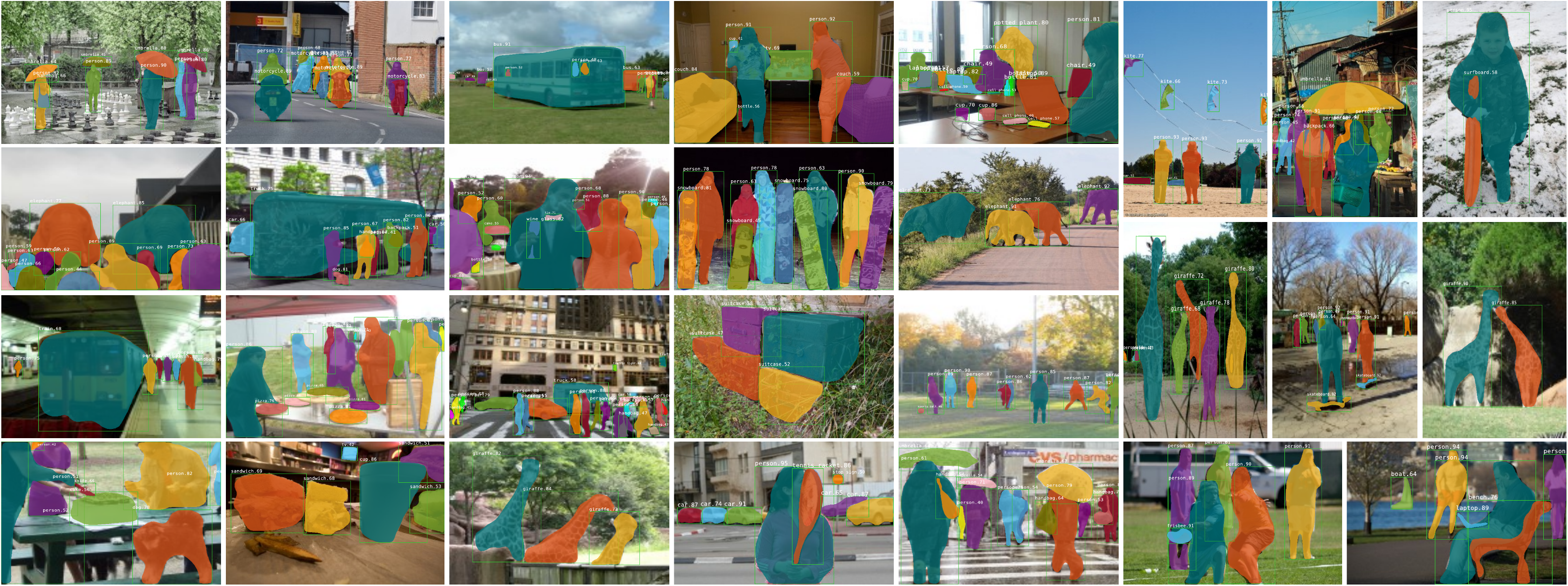}}\vspace{-2.mm}
\caption{
\textbf{Qualitative Results} of ISTR on the COCO \texttt{test-dev} split. Predictions are shown with a threshold of $0.4$
}\vspace{-0.5mm}
\label{fig4}
\end{figure*}
\paragraph{Pooling Type:} In Table~\ref{tab1-e}, we evaluate various strategies for obtaining the image features by extracting different information from input images.
%
%
As can be seen, the global max-pooling does not obtain a high score, while the global average pooling performs better.
We believe this is because the max-pooling extracts features from the highest activated pixel in the feature map, which usually corresponds to a single activated object.
In contrast, the global average pooling extracts features that contain information about the whole image.
We also find that the position embeddings are essential for achieving high results.
By summing the position embeddings with averaged image features, the performance is significantly improved.
\subsection{Stage Analysis}
One of the essential components of ISTR is the recurrent refinement strategy, which provides a new way to achieve instance segmentation compared to the bottom-up and top-down strategies.
The prediction heads infer the classes, bounding boxes, and masks for each instance using the query boxes updated in each stage.
%
%
We investigate the performance of the recurrent refinement stages both quantitatively and qualitatively in Fig.~\ref{fig3}.
%
From the visualization of masks and bounding boxes shown in Fig.~\ref{fig3-3}, we can see that both masks and boxes are refined from coarse to fine.
The mask and box APs shown in Fig.~\ref{fig3-1} and Fig.~\ref{fig3-2} using different backbones also demonstrate that the results are gradually refined step by step.
\subsection{Main Results}
We compare ISTR with the state-of-the-art instance segmentation methods as well as the latest end-to-end object detection methods to demonstrate its superior performance.
%
\paragraph{Quantitative Results:} From Table~\ref{tab2}, we can see that ISTR performs well, especially on small objects.
For example, the AP$^m_S$ of ISTR based on ResNet101-FPN outperforms SOLOv2 based on ResNet101-FPN by $5.5$ points.
We believe this is because the bipartite matching cost does not filter small objects for training.
MEInst also uses mask embeddings for instance segmentation.
However, the performance of MEInst suffers significantly due to the redundant predictions of mask embeddings.
For example, the AP$^m$ of ISTR based on ResNet101-FPN outperforms MEInst based on ResNet101-FPN by a large margin of $4.6$ points.
Besides, we also find performance gains of ISTR in detection when comparing the results with the state-of-the-art end-to-end object detection methods.
We find that the AP$^b$ of ISTR outperforms DETR and sparse R-CNN based on ResNet101-FPN by $4.6$ and $2.5$ points, respectively.
Overall, it is surprising that, despite the suboptimal mask embeddings from PCA, ISTR can still obtain such a good result.
This demonstrates the strength of the proposed end-to-end mechanism and shows the potential of concurrently conducting detection and segmentation with Transformers.
\paragraph{Qualitative Results:} We show some examples comparing ISTR with Mask R-CNN in Fig.~\ref{fig4-1}.
As can be seen, Mask R-CNN suffers inferior performance when NMS does not remove the duplicate predictions.
More visualization results in Fig.~\ref{fig4-2} suggest that, although ISTR obtains state-of-the-art mask APs, there is still room for further improvement by learning finer masks.
We leave this for future work.

\section{Conclusion}
In this paper, we propose a new framework, termed instance segmentation Transformer (ISTR), to explore the end-to-end mechanism for instance segmentation.
ISTR predicts low-dimensional mask embeddings instead of high-dimensional masks, which inspires the design of a mask matching cost and facilitates the regression.
Besides, ISTR concurrently conducts detection and segmentation via a recurrent refinement strategy, which provides a new perspective to achieve end-to-end instance segmentation and boosts the performance of both tasks.
On the challenging COCO dataset, the strong performance of ISTR demonstrates its potential for instance-level recognition tasks.

{\small
\bibliographystyle{ieee_fullname}
\bibliography{egbib}
}

\end{document}